\documentclass[submission,copyright,creativecommons]{eptcs}
\providecommand{\event}{TARK 2023} 

\usepackage{graphicx,float,caption}
\usepackage{iftex}
\usepackage{amsmath}
\usepackage{amssymb}
\usepackage{array,multirow,makecell}
\usepackage{diagbox}
\usepackage{color}
\usepackage{colortbl}
\setcellgapes{1pt}
\makegapedcells
\newcolumntype{R}[1]{>{\raggedleft\arraybackslash }b{#1}}
\newcolumntype{L}[1]{>{\raggedright\arraybackslash }b{#1}}
\newcolumntype{C}[1]{>{\centering\arraybackslash }b{#1}}

\usepackage{biblatex}
\title{Mining for Unknown Unknowns}
\author{Bernard Sinclair-Desgagné
\institute{SKEMA Business School\\ Sophia Antipolis, France}
\institute{GREDEG, Université Côte d'Azur\thanks{I am grateful to the participants at several seminars for constructive questions and remarks. I also owe special thanks to Thierry Bréchet, Thierry Burger-Helchem, Michel Desmarais, Soumittra Dutta, Ludovic Dibiaggio, Armand Hatchuel, Emile Grenier-Robillard, Henry Tulkens, and Benoit Weil for their encouragements, comments or suggestions. Three diligent and competent reviewers contributed to substantially clarify and improve the initial draft of this paper.}\\
Sophia Antipolis, France}
\email{bsd@skema.edu}}

\def\titlerunning{Mining for Unkown Unkowns}
\def\authorrunning{B. Sinclair-Desgagné}
\begin{document}
\maketitle

\begin{abstract}
Unknown unknowns are future relevant contingencies that lack an ex ante description. While there are numerous retrospective accounts showing that significant gains or losses might have been achieved or avoided had such contingencies been previously uncovered, getting hold of unknown unknowns still remains elusive, both in practice and conceptually. Using Formal Concept Analysis (FCA) - a subfield of lattice theory which is increasingly applied for mining and organizing data - this paper introduces a simple framework to systematically think out of the box and direct the search for unknown unknowns.
\end{abstract}
\bigskip

\begin{quote}
{\footnotesize There are only two kinds of campaign plans, good ones and bad
ones.\\The good ones almost always fail through unforeseen
circumstances\\that often make the bad ones succeed.}

{\footnotesize \ \ \ \ \ \ \ \ \ \ \ \ \ \ \ \ \ \ \ \ \ \ \ \ \ \ \ \ \ \ \
\ \ \ \ \ \ \ \ \ \ \ \ \ \ \ \ \ \ \ \ \ \ \ \ \ \ - Napoleon Bonaparte
-\bigskip }
\end{quote}


\section{Introduction}

As the recent Covid-19 pandemic reminded us, life is filled with unknown
unknowns -- i.e. contingencies one cannot be aware of ex ante, much less fit
into standard risk analysis. In addition to a wealth of examples coming from
history and politics, unknown unknowns are now well-documented, and their
importance is acknowledged, in many areas of economics and management such
as public policy \parencite{Mueller2020}, business strategy \parencite{Bryan2021,Ehrig2022a}, entrepreneurship \parencite{Ehrig2022b}, contracts and the theory of the firm \parencite{Tirole1999}, and security \parencite{Rashid2016}.

To be sure, getting hold of such contingencies might allow to achieve
significant payoffs or avoid major losses. Substantial research efforts have
thus been expended, and notable advances been made, in this direction. To
get a rigorous conceptual grasp at the notion of unknown unknowns, one may
now draw, notably, from the literatures on Knightian uncertainty (e.g.,
\cite{Bewley1998}), undescribable events (e.g., \cite{Al-Najjar2006}), unforeseen
contingencies (e.g., \cite{Dekel1998, Kochov2018}), unawareness (e.g.,
\cite{Schipper2014a,Schipper2014b}), and surprises \parencite{Taleb2007,Palm2012}. Yet, for
someone who would primarily want to uncover ahead of time the\ concrete
unknown unknowns she might be facing, the task would remain elusive.

This paper will now seek to meet this demand. Similar endeavors have already
been tried, and results obtained, in areas where unknown unknowns occur
frequently: like C-K Theory \parencite{Hatchuel2009}, TRIZ \parencite{Ilevbare2013}, and creativity support systems \parencite{Gabriel2016, Wang2017} in innovation management; knowledge spaces \parencite{Doignon1999} in learning; and elicitation methods \parencite{Lakkaraju2017,Ramasesh2014,Sutcliffe2013} in engineering. As it turns out,
Formal Concept Analysis (FCA), which I will be using here, provides an
appropriate language, and especially structure, to build a framework which
is both rigorous (it is grounded in lattice theory) and operational (its
implementation requires only spreadsheets).

The suggested scheme is sketched in the following Section 2. It is next
developed rigorously and with more generality in Section 3. A fourth section
contains concluding remarks.\bigskip


\section{An informal account}

Consider a 3x3 matrix with horizontal coordinates A, B, C, and vertical
coordinates $\alpha $ $\beta $, $\gamma $. For concreteness, the former
might refer to different objects, items or events and the latter to various
attributes, characteristics, properties or features. Table 1 shows the features respectively
held by each specific item: object A, for instance, possesses attributes $\alpha $, $\beta $.\smallskip


\begin{table}[h]
\centering
\begin{tabular}{|c|*{3}{c|}}
    \hline 
    \diagbox{Objects}{Attributes} & $\alpha$ & $\beta$ & $\gamma$\\
    \hline A & $\times$ & $\times$ & \\
    \hline B &&& $\times$ \\
    \hline C && $\times$ & $\times$ \\
    \hline 
\end{tabular}
\caption{The existing context}
\end{table}

In Formal Concept Analysis (FCA), such a matrix showing relationships
between `objects' (items, events, etc.) and `attributes' (properties,
features, etc.) is called a context.

In practice, FCA users would of course face much more complicated types of
contexts, with tables comprising dozens of rows and columns, mitigated
relationships between objects and attributes, and (what is crucial for
decision-making) value-weighted properties. But this simple example will
suffice to convey our main points.

The upshot of a discovery, experiment or invention would be the expanded
matrix displayed in Table 2. Two new items -- D and E -- and an extra
characteristic $\delta $ were found. The initial objects A and B now bear
attributes $\delta $ and $\alpha $ respectively, while D possesses
properties $\alpha $ and $\delta $, E exhibits features $\beta $ and $\gamma 
$. This matrix forms a context as well.\smallskip


\begin{table}[h]
\centering
\begin{tabular}{|c|*{4}{c|}}
    \hline 
    \diagbox{Objects}{Attributes} & $\alpha$ & $\beta$ & $\gamma$ & \textcolor{red}{$\delta$}\\
    \hline A & $\times$ & $\times$ && \textcolor{red}{$\times$}\\
    \hline B & \textcolor{red}{$\times$} && $\times$ & \\
    \hline C && $\times$ & $\times$ & \\
    \hline \textcolor{red}{D} & \textcolor{red}{$\times$} &&& \textcolor{red}{$\times$}\\
    \hline \textcolor{red}{E} && \textcolor{red}{$\times$} & \textcolor{red}{$\times$} & \\
    \hline
\end{tabular}
\caption{The new context}
\end{table}

Back in time, the incremental component of Table 2 -- i.e. the rows D, E, column $\delta $ and the small x's -- might have been impossible to
describe, much less to anticipate even as random outcomes. They were, so to speak, \textit{unknown unknowns}. Altogether, they make the context displayed in Table 3.\smallskip


\begin{table}[h]
\centering
\begin{tabular}{|c|*{4}{c|}}
    \hline 
    \diagbox{Objects}{Attributes} & $\alpha$ & $\beta$ & $\gamma$ & \textcolor{red}{$\delta$}\\
    \hline A &&&& \textcolor{red}{$\times$}\\
    \hline B & \textcolor{red}{$\times$} &&& \\
    \hline C &&&&\\
    \hline \textcolor{red}{D} & \textcolor{red}{$\times$} &&& \textcolor{red}{$\times$}\\
    \hline \textcolor{red}{E} && \textcolor{red}{$\times$} & \textcolor{red}{$\times$} & \\
    \hline
\end{tabular}
\caption{The discovery context}
\end{table}

Let us now see how someone could have a grasp at Table 3 using known knowns 
\textbf{only}, these known knowns being the data available from Table 1.

In FCA, the primary mode of organizing the data of a context is
through the use of `concepts'. A concept is defined as a list of objects and
attributes such that the mentioned objects are precisely the ones that share
the listed attributes, and the mentioned attributes are precisely the ones
shared by all the listed objects. Examples of concepts in Table 3 are the
objects B, D with their common attribute $\alpha $, event E with features $\beta $, $\gamma $, items A, D with the shared property $\delta $, and
object D with attributes $\alpha $, $\delta $.

FCA calls an incompletely specified concept, i.e. a list that misses some
object and attributes, a preconcept. The list (B;$\alpha $) is a preconcept
of the concept (B,D;$\alpha $), for instance. Since its object B and
attribute $\alpha $ could already be seen in the existing context of Table 1,
I shall refer to such specific preconcept as a \textit{seed}.

Now, the relationship between B and $\alpha $ is captured in Table 4, which
is actually the negative picture of Table 1. This table constitutes a
context as well, and \textit{it is made only of data from Table 1}. Its
concepts -- which might be called `anti-concepts', since they are the
counterpart of the existing initial concepts -- include (B,C;$\alpha $), (A;$\gamma $) and (B;$\alpha $,$\beta $).\smallskip 


\begin{table}[h]
\centering
\begin{tabular}{|c|*{3}{c|}}
    \hline 
    \diagbox{Objects}{Attributes} & $\alpha$ & $\beta$ & $\gamma$\\
    \hline A &&& Z\\
    \hline B & Z & Z &\\
    \hline C & Z &&\\
    \hline 
\end{tabular}
\caption{The negative existing context}
\end{table}

This paper's main result is that \textit{a seed -} like (B;$\alpha $)\textit{\ - will always be the pre-concept of some anti-concept} -- namely, here,
(B,C;$\alpha $) or (B;$\alpha $,$\beta $). This fact has at least three
ramifications.

First, although one cannot say anything about which objects or which
attributes will be discovered, the structure of the existing context bears
some implications for the structure among discovered objects and attributes.\footnote{
I am grateful to an anonymous referee for this observation.} This opens the
door for establishing a systematic procedure to get some grasp at, and
eventually uncover, unknown unknowns:

(i) Build the negative picture of the existing context;

(ii) Examine the preconcepts of each anticoncept;

(iii) If a seed is found, dig into it to uncover some concepts in the
unknown unknown.\medskip

Second, the latter procedure might be seen as an instance of abduction, a
mode of reasoning associated with creativity and the generation of ideas \parencite{Mabsout2015,Tohme2013}. Unlike deduction, which draws the
logical ramifications of previously given assertions, or induction, which
infers general laws from the observation of recurrent facts, abduction looks
for the best justification (which is here a concept of the discovery
context) after hitting a singular event (a seed).

Third, as the upcoming section will show, it allows for the deployment of a
potentially powerful tool for data exploration and exploitation - namely, Galois
connections. \bigskip


\section{Formal Developments}

Let's now present the mathematics which underlie the scheme outlined above. Subsection 3.1 revisits the basics of Formal Concept Analysis (FCA). Subsection 3.2 next introduces the notion of `revelation mappings'. Subsection 3.3, finally, develops the systematic procedure for thinking out of the box. The treatment is meant to be self-contained. Only set-theoretic arguments are used throughout.\bigskip


\subsection{Basic FCA notions}

A \textit{formal context} is referred to as a triplet $K=(G,M;R)$, where $G$
is a set of \textit{objects}, $M$ a set of \textit{attributes} these objects
may have, and $R$ is a \textit{relation} between $G$ and $M$, i.e. a subset
of the Cartesian product $G\times M$ with the interpretation that $(g,m)\in
R $, or $gRm$, if object $g$ has attribute $m$.

Denote $\wp (G)$ and $\wp (M)$ the respective power sets (or sets of all
subsets) of $G$ and $M$. Set inclusion $\subseteq $ provides a partial order
on the elements of these sets.\footnote{A set $Q$ is a \textit{partially ordered set} (or \textit{poset}) if there is a relation $\leq $ on $Q$ (called a \textit{partial order}) such that: (i) for $q\in Q$, $q\leq q$ (reflexivity property); (ii) for $q_{1}$, $q_{2}\in Q$, $q_{1}\leq q_{2}$ and $q_{2}\leq q_{1}$ implies $q_{1}=q_{2}$ (antisymmetry); for $q_{1}$, $q_{2}$, $q_{3}\in Q$, $q_{1}\leq q_{2}$ and $q_{2}\leq q_{3}$ implies $q_{1}\leq q_{3}$ (transitivity).
\par
{}} The following set-to-set functions $I_{R}$ and $E_{R}$ defined as
\begin{eqnarray*}
    \text{for \ S} &\subseteq &G\text{, \ }I_{R}(S)=\left\{ m\in M:gRm\text{ \ for\ all \ g}\in S\right\}  \\
    \text{for \ T} &\subseteq &M\text{, \ }E_{R}(T)=\left\{ g\in G:gRm\text{ \ for\ all \ m}\in T\right\} 
\end{eqnarray*}

{\footnotesize \noindent }are called the \textit{Birkhoff Operators} for $G$
and $M$ respectively. For a set of objects $S$, $I_{R}(S)$ - the \textit{intent} of $S$ - gives all the attributes in $T$ which these objects have in
common. For a given set of attributes $T$, $E_{R}(T)$ - the \textit{extent }of $T$ - gives all the objects in $S$ that share these attributes. In the
context displayed in Table 1, $I_{R}($A,C$)=\{\beta \}$ and $E_{R}(\alpha
,\beta )=\{$A$\}$.

A well-known property of the Birkhoff Operators is that of \textit{duality}:
knowing $I_{R}(\cdot )$ completely determines $E_{R}(\cdot )$, and
vice-versa, specifying $E_{R}(\cdot )$ also defines $I_{R}(\cdot )$.

A \textit{formal concept} in the context $K=(G,M;R)$ is now a pair of sets $(Q;V)$, with $Q\subseteq G$ and $V\subseteq M$, such that $I_{R}(Q)=V$ and $E_{R}(V)=Q$. The extent of a concept $(Q;V)$ is thus $Q$, while its intent
is $V$.\footnote{The way FCA defines a formal concept agrees with the International Standard Organization's ISO 704 definition: \textquotedblleft In a concept, one distinguishes its `intension' and `extension'. The intension of a concept comprises all attributes thought with it, the extension comprises
all objects for which the concept can be predicated. In general, the richer the intension of a concept is, the lesser is its extension, and vice versa.\textquotedblright}

A \textit{preconcept} in $K$, finally, is a pair $(P;U)$, with $P\subseteq G$
and $U\subseteq M$, such that $P\subseteq E_{R}(U)$ or, equivalently, $U\subseteq I_{R}(P)$. Preconcepts can be ordered as follows \parencite{Denniston2013}: $(P;U)\sqsubseteq (P^{\prime };U^{\prime })$, meaning
that $(P;U)$ is \textit{less extensive than} $(P^{\prime };U^{\prime })$, if 
$P\subseteq P^{\prime }$ and $U\subseteq U^{\prime }$.\bigskip


\subsection{Revelation mappings}

From now on, $K=(G,M;R)$ will denote the existing context and $K^{+}=(G^{+},M^{+};R^{+})$ the new context after the previous unknown
unknowns have been revealed.\footnote{Power sets, Birkhoff Operators, formal concepts, and preconcepts are similarly defined on their context of reference, be it $K$, $K^{+}$, or any other context.}

Let's then call \textit{revelation mappings} the functions $\Phi :\wp (G^{+})\rightarrow \wp (M^{+})$, $\Psi :\wp (M^{+})\rightarrow \wp (G^{+})$ such that\footnote{Let's agree that $I_{R}(g)= \varnothing $ when $g\notin G$, and $E_{R}(m)= \varnothing $ when $m\notin M$.}
\begin{eqnarray*}
    \text{for\ S}^{+} &\subseteq &G^{+}\text{, \ }\Phi (S^{+})=I_{R^{+}}(S^{+})\text{ }\setminus \text{ }\underset{g\in S^{+}}{\cup }I_{R}(g) \\
    \text{for \ T}^{+} &\subseteq &M^{+}\text{, \ }\Psi (T^{+})=E_{R^{+}}(T^{+})\text{ }\setminus \text{ }\underset{m\in T^{+}}{\cup }E_{R}(m)
\end{eqnarray*}

\noindent If one takes a set $S\subseteq G$ of objects from the existing
context $K$, $\Phi (S)$ delivers the set of attributes (old or new) in $M^{+}
$ which are newly associated with these objects. In Table 2, for instance, $\Phi ($B,E$)=\varnothing $ and $\Phi ($A$)=\{{\Large \delta }\}$. Similarly,
for a subset of initial attributes $T\subseteq M$, $\Psi (T)$ gives all (and
only) the initial or new objects that now possess these attributes. For
example, $\Psi (\gamma ,\delta )=\varnothing $ but $\Psi (\beta ,\gamma )=\{$E$\}$.

As for the Birkhoff operators, there is a \textit{duality property} between $\Phi (\cdot )$ and $\Psi (\cdot )$: each one uniquely characterizes the
other. These mappings also hold additional features which are spelled out in
the upcoming propositions.

First, say that a function $\pi :X\rightarrow Y$ between two sets $X$ and $Y$, partially ordered by $\leq $ and $\preceq $ respectively, is \textit{antitone} (or order-reversing) if, for $p_{1},p_{2}\in X$, $p_{1}\leq p_{2}$
\ implies $\ \pi (p_{2})\preceq \pi (p_{1})$. A first statement is now at
hand.\medskip

{\footnotesize \noindent }\textsc{Proposition 1: }The revelation mappings $\Phi $ and $\Psi $ are antitone.\medskip

{\footnotesize \noindent }\textsc{Proof:}

First, consider $\Phi $. Take two sets $S_{1}^{+}$, $S_{2}^{+}\in \wp
(G^{+}) $ such that $S_{1}^{+}\subseteq S_{2}^{+}$; we must show that $\Phi
(S_{2}^{+})\subseteq \Phi (S_{1}^{+})$. If $m\in \Phi (S_{2}^{+})$, then $m\in $ $I_{R^{+}}(S_{2}^{+})$ so $gR^{+}m$ for all $g\in S_{2}^{+}$. Since $S_{1}^{+}\subseteq S_{2}^{+}$, we have that $gR^{+}m$ for all $g\in
S_{1}^{+} $, hence $m\in $ $I_{R^{+}}(S_{1}^{+})$. Now, if $m\notin M$, $m\notin I_{R}(g)$ for any $g\in S_{1}^{+}$; it follows that $m\in
I_{R^{+}}(S_{1}^{+})$ $\setminus $ $\underset{g\in S_{1}^{+}}{\cup }I_{R}(g)=\Phi (S_{1}^{+})$. Suppose, alternatively, that $m\in M$. Since $m\in \Phi (S_{2}^{+})$, it must be the case that $not(gRm)$ for all $g\in
S_{2}^{+}$, hence $not(gRm)$ as well for all $g\in S_{1}^{+}$ since $S_{1}^{+}\subseteq S_{2}^{+}$; it follows again that $m\in \Phi (S_{1}^{+})$. This shows that $\Phi (S_{2}^{+})\subseteq \Phi (S_{1}^{+})$. The same
line of reasoning works for $\Psi $ (as can be expected from duality).\textsc{\ \ \ \ \ \ \ \ \ \ \ \ \ \ \ \ \ \ \ \ \ \ \ \ \ \ \ \ \ \ \ \ \ \
\ \ \ \ \ \ \ \ \ \ \ \ \ \ \ \ \ \ \ \ \ \ \ \ \ \ \ \ \ \ \ \ \ \ \ \ \ \ \ \ \ \ \ \ \ \ \ \ \ \ \ \ \ \ \ \ \ \ \ \ \ \ \ \ \ \ \ \ \ \ \ \ \ \ \ \ \ \ \ \ \ \ \ \ \ \ \ \ \ \ \ \ \ \  }$\blacksquare $\textsc{\medskip }

This property of revelation mappings means that, the more objects or
attributes one starts with, the more demanding it is to find new
relationships that fit them all. This intuitive result is also instrumental
in deriving other important characteristics of revelation mappings.

A key notion to introduce at this point is that of a \textit{Galois
connection}.\footnote{Since at least Ore (1944)'s seminal article \parencite{Ore1944}, Galois connections have been increasingly employed throughout mathematics and computer science. To go beyond the very short primer offered in this paper, the reader may look at \parencite{Davey2002, Doignon1999,Ganter1999}, and some of their common references.} Let $X$ and $Y$ be two sets
partially ordered by $\leq $ and $\preceq $ respectively. A (antitone) 
\textit{Galois connection} $(\pi ,\theta )$ on $X$ and $Y$ is a pair of
functions $\pi :X\rightarrow Y$ and $\theta :Y\rightarrow X$ such that the
following equivalent properties are satisfied.

{\footnotesize \noindent }(i) \ For each $p\in X$, $p\leq \theta \pi (p)$
and for each $q\in Y$, $q\preceq \pi \theta (q)$.

{\footnotesize \noindent }(ii) For $p\in X$ and $q\in Y$, $\ p\leq \theta
(q) $ \ if and only if $q\preceq \pi (p)$.\medskip

It is well-known that the Birkhoff operators $(I_{R},E_{R})$, $(I_{R^{+}},E_{R^{+}})$ are antitone Galois connections on, respectively, the power sets $\wp (G)$, $\wp (M)$ and $\wp (G^{+})$, $\wp (M^{+})$ ordered by set inclusion (see, e.g., \cite{Ganter1999}, p. 13-14).\ In this case,
property (i) means that the attributes common to a given set of objects
might be shared by more objects, while the objects that share a given set of
attributes might have more attributes in common. Property (ii), on the other
hand, says that some objects are among those sharing a given set of
attributes if and only if these attributes are among those common to these
objects.

As it turns out, the pair of revelation mappings $(\Phi ,\Psi )$ forms a
Galois connection.\medskip

{\footnotesize \noindent }\textsc{Proposition 2: }The pair of revelation
mappings $(\Phi ,\Psi )$ is a Galois connection on the power sets $\wp
(G^{+})$ and $\wp (M^{+})$ partially ordered by inclusion.\medskip

{\footnotesize \noindent }\textsc{Proof:}

To see this, take two sets $S^{+}\in \wp (G^{+})$ and $T^{+}\in \wp (M^{+})$, and notice that
\begin{eqnarray*}
    S^{+} &\subseteq &\Psi (T^{+}) \\
    \text{if and only if \ \ }\forall g &\in &S^{+}\text{, }\forall m\in T^{+}\text{: \ }gR^{+}m\text{ \ and \ }not(gRm) \\
    \text{if and only if \ }\forall m &\in &T^{+}\text{, }\forall g\in S^{+}\text{: \ }gR^{+}m\text{ \ and \ }not(gRm) \\
    \text{if and only if \ \ }T^{+} &\subseteq &\Phi (S^{+})
\end{eqnarray*}
    \begin{flushright}
        $\blacksquare$
    \end{flushright}    
Proposition 2 underlies a central result. Like any Galois connection (\cite{Ganter1999}, p. 14), $(\Phi ,\Psi )$ establishes a relation, noted $R_{(\Phi ,\Psi )}^{+}$, between the set of objects $G^{+}$ and the set of attributes $M^{+}$. This relation is defined as
\begin{eqnarray*}
    R_{(\Phi ,\Psi )}^{+} &=&\left\{ (g,m)\in G^{+}\times M^{+}\mid g\in \Psi
    (m)\right\} \\
    &=&\left\{ (g,m)\in G^{+}\times M^{+}\mid m\in \Phi (g)\right\}
\end{eqnarray*}

{\footnotesize \noindent }We can show that $R_{(\Phi ,\Psi )}^{+}$ coincides
with $R^{+}\setminus R$, the set of all new relationships.\medskip

{\footnotesize \noindent }\textsc{Proposition 3: } $R_{(\Phi ,\Psi
)}^{+}=R^{+}\setminus R$ .\medskip

{\footnotesize \noindent }\textsc{Proof: }Observe that $(g,m)\in R_{(\Phi
,\Psi )}^{+}$ \ if and only if \ \ $gR^{+}m$ and $not(gRm)$ ,\ if and only
if\ $\ (g,m)\in R^{+}\setminus R$. 
    \begin{flushright}
        $\blacksquare$
    \end{flushright}
\bigskip

\subsection{Thinking out of the box}

From now on, let $R_{(\Phi ,\Psi )}^{+}=R^{+}\setminus R$ be referred to as $R^{\ast }$. The latter relation defines another formal context, the \textit{discovery} context noted $K^{\ast }=(G^{+},M^{+};R^{\ast })$, which is the context of the unknown unknowns. Can $K^{\ast }$ be inferred from $K$, at least partly? We will now see that the answer actually errs on the yes side.

The ordered pair $(X;Y)$ with $X\neq \varnothing $, $Y\neq \varnothing $ is
called a \textit{seed }in $K$ for $K^{\ast }$ if it is a preconcept in $K^{\ast }$ while $X\subseteq G$ and $Y\subseteq M$. As the next statement
confirms, the existence of a seed is guaranteed when the existing context
harbors at least one new relationship between the original objects and
attributes.\medskip

{\footnotesize \noindent }\textsc{Proposition 4: }If $R^{\ast }\cap (G\times
M)\neq \varnothing $, then there is at least one seed in $K$ for $K^{\ast }$.\medskip

{\footnotesize \noindent }\textsc{Proof:}

The assumption implies that there is at least one concept $(Q;V)$ in $K^{\ast }$ such that $Q\cap G\neq \varnothing $ and $V\cap M\neq \varnothing 
$. Since $Q\cap G\subseteq Q=I_{R^{\ast }}(V)\subseteq I_{R^{\ast }}(V\cap
M) $ and $V\cap M\subseteq V=E_{R^{\ast }}(Q)\subseteq E_{R^{\ast }}(Q\cap
G) $, the pair $(Q\cap G;V\cap M)$ is a preconcept in $K^{\ast }$. \ \ \ $\
\ \ \ \ \ \ \ \ \ \ \ \ \ \ \ \ \ \ \ \ \ \ \ \ \ \ \ \ \ \ \ \ \ \ \ \ \ \ \ \ \ \ \ \ \ \ \ \ \ \ \ \ \ \ \ \ \ \ \ \ \ \ \ \ \ \ \ \ \ \ \ \ \ \ \ \ \ \ \ \ \ \ \ \ \ \ \ \ \ \ \ \ \ \ \ \ \ \ \ \ \blacksquare $\medskip

As suggested in Section 2, looking for seeds might be a reasonable first
step to uncover unknown unknowns. The major reason is that, as we will now
demonstrate, \textit{it is possible to characterize the location of seeds}.

First, according to the following proposition, a seed must combine objects
and attributes which are a priori unrelated.\medskip

{\footnotesize \noindent }\textsc{Proposition 5: }No preconcept (a fortiori
concept) in the existing context $K$ can be a seed for the discovery context 
$K^{\ast }$.\medskip 

{\footnotesize \noindent }\textsc{Proof:}

Let $(P;U)$ be a preconcept in $K$. By definition, $\Phi (P)=I_{R^{+}}(P)$ 
$\setminus $ $\underset{g\in P}{\cup }I_{R}(g)$. But $U\subseteq I_{R}(P)=\underset{g\in P}{\cap }I_{R}(g)\subseteq \underset{g\in P}{\cup }I_{R}(g)$.
It follows that $U\nsubseteq \Phi (P)$, hence $(P;U)$ is not a preconcept in 
$K^{\ast }. \text{\ \ \ \ \ \ \ \ \ \ \ \ \ \ \ \ \ \ \ \ \ \ \ \ \ \ \ \ \ \ \ \ \ \ \ \ \ \ \ \ \ \ \ \ \ \ \ \ \ \ \ \ \ \ \ \ \ \ \ \ \ \ \ \ \ \ \ \ \ \ \ \ \ \ \ \ \ \ \ \ \ \ \ \ \ \ \ \ \ \ \ \ \ \ \ \ \ \ \ \ \ \ \ \ \ \ \ \ \ \ \ \ \ \ \ \ \ \ \ \ \ \ \ \ \ \ \ \ \ \ \ \ \ \ \ \ \ \ \ \ \ \ \ \ \ \ \ \ \ \ \ \ \ \ \ \ \ \ \ \ \ \ \ \ \ \ \ \ \ \ \ \ \ \ \ \ \ \ \ \ \ \ \ \ \ \ \ \ \ \ \ \ \ \ \ \ \ \ \ \ \ \ \ \ \ \ \ \ \ \ \ \ \ \ \ \ \ \ \ \ \ \ \ \ \ \ \ \ \ \ \ \ \ \ \ \ \ \ \ \ \ \ \ \ \ \ \ \ \ \ \ \ \ \ \ \ \ \ \ \ \ \ \ \ \ \ \ \ \ \ \ \ \ \ \ \ \ \  }\blacksquare$ \medskip

This result tells us something about how not to look for novelties. A
corollary is that a seed in $K$ for $K^{\ast }$ must be a pair $(P;U)$, with 
$P\subseteq G$ and $U\subseteq M$, such that $P\cap (\underset{m\in U}{\cup }E_{R}(m))=\varnothing $ and $U\cap (\underset{g\in P}{\cup }I_{R}(g))=\varnothing $. This suggests working with the negative of the
existing context $K$, noted $\overline{K}=(G,M;\overline{R})$, where the
relation $\overline{R}=G\times M\setminus R$ refers to the \textit{reverse
relation} $g\overline{R}m$ which holds when object $g$ \textit{does not}
have attribute $m$. The next (key, and somewhat surprising) proposition
shows that $\overline{K}$ - which can be obtained using \textbf{only} the
initial data - is the appropriate `outbox' in which mining for unknown
unknowns might begin.\medskip 

{\footnotesize \noindent }\textsc{Proposition 6: }A seed is a preconcept of
the negative existing context $\overline{K}$.\medskip 

{\footnotesize \noindent }\textsc{Proof:}

Let $(P;U)$ be a seed for $K^{\ast }$. Then $U\subseteq \Phi (P)\cap M=M\cap
I_{R^{+}}(P)\setminus \underset{g\in P}{\cup }I_{R}(g)\subseteq M\setminus 
\underset{g\in P}{\cup }I_{R}(g)\\
=\underset{g\in P}{\cap }(I_{R}(g))^{c}=I_{\overline{R}}(P).$\ \ \ \ \ \ \ \ \ \ \ \ \ \ \ \ \ \ \ \ \ \ \ \ \ \ \ \ \ \ \ \ \ \ \ \ \ \ \ \ \ \ \ \ \ \ \ \ \ \ \ \ \ \ \ \ \ \ \ \ \ \ \ \ \ \ \ \ \ \ \ \ \ \ \ \ \ \ \ \ \ \ \ \ \ \ \ \ \ \ \ \ \ \ \ \ \ \ \ \ \ \ \ \ \ \ \ \ \ \ \ \ \ \ \ \ \ \ \ \ \ \ \ \ \ \ \ \ \ \ \ \ \ \ \ \ \ \ \ \ \ \ \ \ \ \ \ \ \ \ \ \ \ $\blacksquare $\medskip

Seeds for the discovery context $K^{\ast }$ - which comprises a priori
unknown relationships between objects in $G$ and attributes in $M$ - thus
happen to point, not only at concepts in $K^{\ast }$, but also at the
concepts of the negative existing context $\overline{K}$. This suggests the
procedure already outlined in Section 2:

\ \noindent $\bullet $ {\footnotesize \ \ }Take the negative context $\overline{K}$ of $K$;

\ \noindent $\bullet $ \ \ Consider a concept in $\overline{K}$ (i.e. an anti-concept);

\ \noindent $\bullet $ \ \ Examine the latter's preconcepts;

\ \noindent $\bullet $ \ \ If one of these preconcepts brings out a new
relationship between its objects and attributes, then a seed has been found which anticipates some concepts in the discovery context $K^{\ast }$.\medskip

Whether this scheme can be fruitful in practice remains to be seen. One
hurdle could be computational complexity (see the concluding remarks).

Interestingly, however, Propositions 5 and 6 suggest that concepts in the
negative existing context $\overline{K}$ can be constructed using the
mappings $\ \widetilde{\Phi }:\wp (G)\rightarrow \wp (M)$ and $\widetilde{\Psi }:\wp (M)\rightarrow \wp (G)$ defined as
\begin{eqnarray*}
    \text{for \ S} &\subseteq &G\text{, \ }\widetilde{\Phi }(S)=M\text{ }\setminus 
    \text{ }\underset{g\in S}{\cup }I_{R}(g) \\
    \text{for \ T} &\subseteq &M\text{, \ }\widetilde{\Psi }(T)=G\text{ 
 }\setminus 
    \text{ }\underset{m\in T}{\cup }E_{R}(m)
\end{eqnarray*}\

{\footnotesize \noindent }respectively. Comparing the latter expressions
with the ones corresponding to the above revelation mappings, the functions $\widetilde{\Phi }$ and $\widetilde{\Psi }$ can be seen as approximations for 
$\Phi $ and $\Psi $. Whether closer approximations (in a sense to be made
precise) can be found, which would then provide a better grasp at unknown
unknowns, would be a valuable research topic.\bigskip


\section{Concluding remarks}

This paper submitted a new framework and approach to handle unknown
unknowns. The scheme has rigorous foundations in lattice theory. It looks
widely applicable, furthermore, since it can incorporate various kinds of data --
quantitative and qualitative, objective and subjective, financial and
non-financial. And it seems to be user-friendly, boiling down to using only
spreadsheets.

At this stage, in addition to the extensions suggested at the end of the
previous section, other ones could be the following:

First, on a technical note, listing all the concepts of a formal context is
generally burdensome.\footnote{An upper bound on the number of concepts in the context K = (G,M;R) is $\frac{3}{2}2^{\sqrt{\mid R\mid +1}}-1$. See \cite{Ganter1999}, p. 94.} Yet, the search for seeds requires this exercise. Research and development on how to identify concepts in a given context is very much ongoing. Several algorithms and softwares already exist: many (mentioned in \cite{Singh2016}, for instance) are subject to a patent but others -- GALICIA and JALABA, for example -- can be freely downloaded. Two promising trends are to
take full advantage of negative information (i.e. the information contained
in the negative existing context $\overline{K}$), as in \cite{Rodriguez-Jimenez2016}  or \cite{Perez-Gamez2021}, and to assign weights to attributes, as in \cite{Belohlavek2011}.

Second, the above derivation made minimal assumptions about the use of a
priori knowledge, ignoring issues of landscape and timing, and forbidding
the use of probabilities. In practice, however, one might be able to tap on
probabilistic beliefs based on science, predictive models or sound
experience, in order to figure out the plausibility of new relationships
between objects and attributes. This endeavor will enhance the search for
seeds, hence the prospecting for unknown-unknowns.\bigskip

\nocite{*}
\printbibliography
\end{document}